%% file: main.tex
\documentclass[runningheads]{llncs}
\usepackage{graphicx}
\usepackage{siunitx}
\usepackage{color}
\usepackage{amssymb}
\usepackage{amsmath}
\usepackage{makecell}
\usepackage{multirow}
\usepackage{lipsum}
\usepackage{caption}
\usepackage{subcaption}
\usepackage{booktabs}
\usepackage{algorithm} 
\usepackage{algpseudocode} 

\usepackage{pifont}
\newcommand{\sdt}{SDT~}
\newcommand{\sdtt}{SDT}
\newcommand{\bfsection}[1]{\vspace*{0cm}\noindent\textbf{#1.}}

\usepackage{pifont}

\newcommand{\red}[1]{\textcolor{red}{#1}}

\newcommand{\etal}{\textit{et al}.}
\newcommand{\ie}{\textit{i}.\textit{e}.}
\newcommand{\eg}{\textit{e}.\textit{g}.}

\newcommand*{\affaddr}[1]{#1} 
\newcommand*{\affmark}[1][*]{\textsuperscript{#1}}

\usepackage[x11names]{xcolor}
\usepackage[pagebackref=true,breaklinks=true,letterpaper=true,colorlinks=false,bookmarks=false]{hyperref}
\hypersetup{
     colorlinks = true,
     linkcolor = red,
     anchorcolor = black,
     citecolor = SpringGreen4,
     filecolor = black,
     urlcolor = black
     }

\makeatletter
\def\@fnsymbol#1{\ensuremath{\ifcase#1\or \dagger\or *\or \ddagger\or
   \mathsection\or \mathparagraph\or \|\or **\or \dagger\dagger
   \or \ddagger\ddagger \else\@ctrerr\fi}}
\makeatother
     
\begin{document}

\title{Structure-Preserving Instance Segmentation via Skeleton-Aware Distance Transform}
\titlerunning{Skeleton-Aware Distance Transform}

\author{
Zudi Lin\affmark[1]\thanks{\footnotesize  Currently  affiliated with Amazon Alexa. Work was done before joining Amazon.} \and
Donglai Wei\affmark[2] \and
Aarush Gupta\affmark[3]\thanks{\footnotesize Work was done during an internship at Harvard University.} \and
Xingyu Liu\affmark[1]$^{*}$ \and
Deqing Sun\affmark[4] \and
Hanspeter Pfister\affmark[1]
}
\authorrunning{Lin \etal}
\institute{
\affaddr{\affmark[1]Harvard University} \quad
\affaddr{\affmark[2]Boston College} \quad
\affaddr{\affmark[3]CMU} \quad
\affaddr{\affmark[4]Google Research}
}

\maketitle
\input{sections/0_abstract.tex}
\input{sections/1_introduction.tex}
\input{sections/2_related.tex}
\input{sections/3_method.tex}
\input{sections/4_experiment.tex}

\section{Conclusion}

In this paper, we introduce the {\em skeleton-aware} distance transform (\sdtt) to capture both the geometry and topological connectivity of instance masks with complex shapes. For multi-class problems, we can use class-aware semantic segmentation to mask the SDT energy trained for all objects that is agnostic to their classes.
We hope this work can inspire more research on not only better representations of object masks but also novel models that can better predict those representations with shape encoding. 
We will also explore the application of \sdt in the more challenging 3D instance segmentation setting.

\bfsection{Acknowledgments}
This work has been partially supported by NSF awards IIS-2239688 and IIS-2124179. This work has also been partially supported by gift funding and GCP credits from Google.



\bibliographystyle{splncs04}
\bibliography{egbib}

\end{document}

%% file: sections/0_abstract.tex
\begin{abstract}
%
Objects with complex structures pose significant challenges to existing instance segmentation methods that rely on boundary or affinity maps, which are vulnerable to small errors around contacting pixels that cause noticeable connectivity change.
While the distance transform (DT) makes instance interiors and boundaries more distinguishable,  it tends to overlook the intra-object connectivity for instances with varying width and result in over-segmentation.  
To address these challenges, we propose a {\em skeleton-aware} distance transform (\sdtt)  that combines the merits of object skeleton in preserving connectivity and DT in modeling geometric arrangement to represent instances with arbitrary structures. 
%
%
Comprehensive experiments on histopathology image segmentation demonstrate that \sdtt{} achieves state-of-the-art performance. 
%
%
\end{abstract}

%% file: sections/1_introduction.tex
\section{Introduction}\label{sec:intro}
\input{tex-fig/fig_teaser.tex}

Instances with complex shapes arise in many biomedical domains, and their morphology carries critical information. For example, the structure of gland tissues in microscopy images is essential in accessing the pathological stages for cancer diagnosis and treatment. 
These instances, however, are usually closely in touch with each other and have non-convex structures with parts of varying widths (Fig.~\ref{fig:teaser}\red{a}), posing significant challenges for existing segmentation methods. 


In the biomedical domain, most methods~\cite{briggman2009maximin,ronneberger2015u,chen2016dcan,yan2018deep,qu2019improving} first learns intermediate representations and then convert them into masks with standard segmentation algorithms like connected-component labeling and watershed transform.
These representations are not only efficient to predict in one model forward pass but also able to capture object {\bf geometry} (\ie, precise instance boundary), which are hard for top-down methods using low-resolution features for mask generation.
However, existing representations have several restrictions. For example, boundary map is usually learned as a pixel-wise binary classification task, which makes the model conduct relatively local predictions and consequently become vulnerable to small errors that break the connectivity between adjacent instances (Fig.~\ref{fig:teaser}\red{b}). 
To improve the boundary map, Deep Watershed Transform (DWT)~\cite{bai2017deep} predicts the Euclidean distance transform (DT) of each pixel to the instance boundary. This representation is more aware of the structure for convex objects, as the energy value for centers is significantly different from pixels close to the boundary.
However, for objects with non-convex morphology, the boundary-based distance transform produces multiple local optima in the energy landscape (Fig.~\ref{fig:teaser}\red{c}), which tends to break the intra-instance connectivity when applying thresholding and results in over-segmentation.

To preserve the {\bf connectivity} of instances while keeping the precise instance boundary, in this paper, we propose a novel representation named {\em skeleton-aware} distance transform (\sdtt). Our \sdt incorporate object skeleton, a concise and connectivity-preserving representation of object structure, into the traditional boundary-based distance transform (DT) (Fig.~\ref{fig:teaser}\red{d}). 
In quantitative evaluations, we show that our proposed \sdt achieves leading performance on histopathology image segmentation for instances with various sizes and complex structures. Specifically, under the Hausdorff distance for evaluating shape similarity, our approach improves the previous state-of-the-art method by relatively $10.6\%$. 


%% file: tex-fig/fig_teaser.tex
\begin{figure}[t]
    \centering
    \includegraphics[width=0.85\columnwidth]{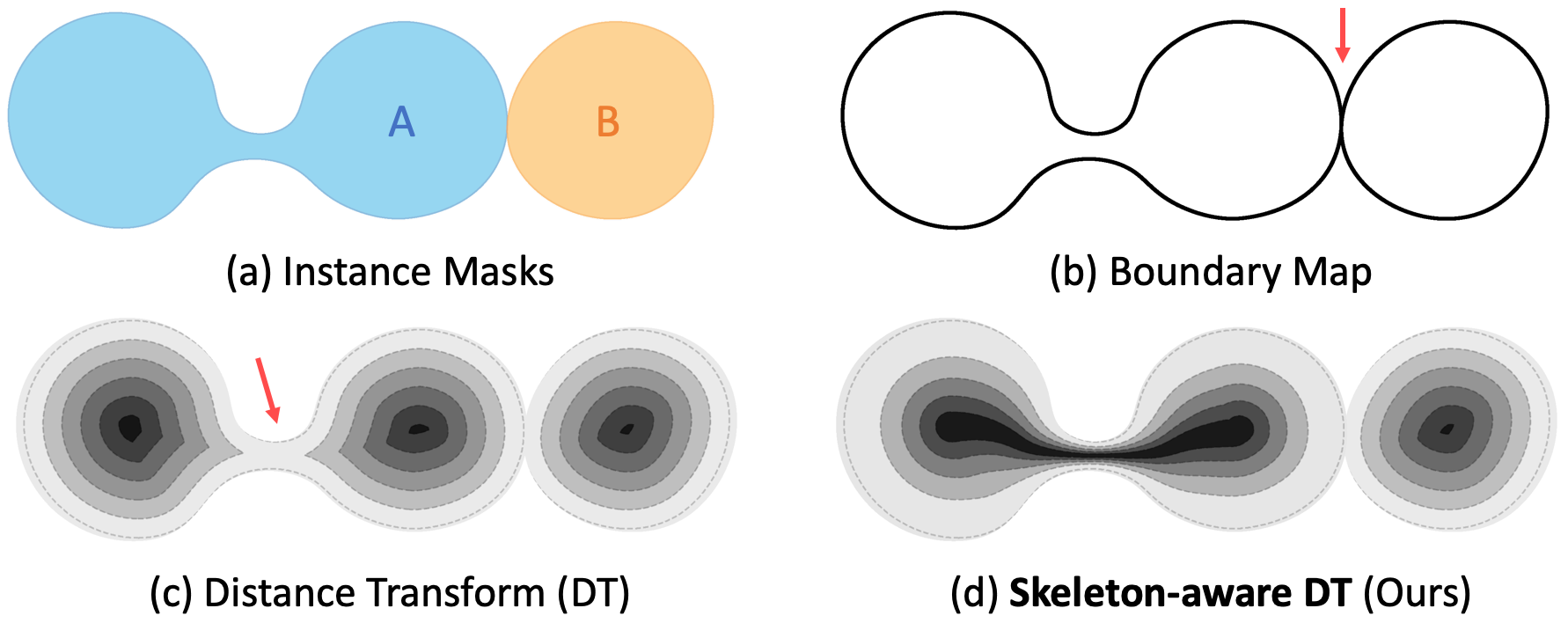}
    \caption{
    Skeleton-aware distance transform (SDT). Given ({\bf a}) instance masks, ({\bf b}) the boundary map is prone to false merge errors at object contact pixels while ({\bf c}) the distance transform (DT) struggles to preserve object connectivity. ({\bf d}) Our \sdt can both separate touching instances and enforce object connectivity.
    }\label{fig:teaser}
\end{figure}

%% file: sections/2_related.tex
\subsection{Related Work}

\bfsection{Instance Segmentation}
Bottom-up instance segmentation approaches have become de facto for many biomedical applications due to the advantage in segmenting objects with arbitrary geometry.  
U-Net~\cite{ronneberger2015u} and DCAN~\cite{chen2016dcan} use fully convolutional models to predict the boundary map of instances. Since the boundary map is not robust to small errors that can significantly change instance structure, shape-preserving loss~\cite{yan2018deep} adds a curve fitting step in the loss function to enforce boundary connectivity. In order to further distinguish closely touching instances, deep watershed transform (DWT)~\cite{bai2017deep} predicts the distance transform (DT) that represents each pixel as its distance to the closest boundary. However, for complex structure with parts of varying width, the boundary-based DT tends to produce relatively low values for thin connections and consequently causes over-segmentation. 
Compared to DWT, our \sdt incorporates object skeleton (also known as medial axis)~\cite{blum1967transformation,zhang1984fast,lee1994building} that concisely captures the topological connectivity into standard DT to enforce both the geometry and connectivity. 
%

\bfsection{Object Skeletonization}
Object skeleton~\cite{saha2016survey} is a one-pixel wide representation of object masks that can be calculated by topological thinning~\cite{zhang1984fast,lee1994building,Nemeth_Kardos_Palagyi_2011} or medial axis transform~\cite{blum1967transformation}. The vision community has been working on direct object skeletonization from images~\cite{shen2017deepskeleton,ke2017srn,liu2018linear,wang2019deepflux}. Among the works, only Shen \etal~\cite{shen2017deepskeleton} shows the application of the skeleton on segmenting single-object images. We instead focus on the more challenging instance segmentation task with multiple objects closely touching each other. 
Object skeletons are also used to correct errors in pre-computed segmentation masks~\cite{matejek2019biologicallyconstrained}. Our \sdt framework instead use the skeleton in the direct segmentation from images.


%% file: sections/3_method.tex
\section{Skeleton-Aware Distance Transform}

\subsection{\sdt Energy Function}\label{sec:skel_energy}
Given an image, we aim to design a new representation $E$ for a model to learn, which is later decoded into instances with simple post-processing. Specifically, a good representation for capturing complex-structure masks should have two desired properties: {\em precise geometric boundary} and {\em robust topological connectivity}. 

Let $\Omega$ denote an instance mask, and $\Gamma_b$ be the boundary of the instance (pixels with other object indices in a small local neighborhood). The {\em boundary} (or affinity) map is a binary representation where $E|_{\Gamma_b}=0$ and $E|_{\Omega\setminus\Gamma_b}=1.$ 
Taking the merits of DT in modeling the geometric arrangement and skeleton in preserving connectivity, we propose a new representation $E$ that satisfies:
\begin{equation}\label{eqn:property}
    0=E|_{\Gamma_b}<E|_{\Omega\setminus(\Gamma_b\cup\Gamma_s)}<E|_{\Gamma_s}=1
\end{equation}
Here $E|_{\Omega\setminus\Gamma_s}<E|_{\Gamma_s}=1$ indicates that there is only one global maximum for each instance, and the value is assigned to a pixel if and only if the pixel is on the skeleton. This property avoids ambiguity in defining the object interior and preserve connectivity. Besides, $E|_{\Omega\setminus\Gamma_b}>E|_{\Gamma_b}=0$ ensures that boundary is distinguishable as the standard DT, which produces precise geometric boundary.

For the realization of $E$, let $x$ be a pixel in the input image, and $d$ be the metric, \eg, Euclidean distance. The energy function for distance transform (DT) is defined as $E_{\text{DT}}(x)=d(x,\Gamma_b)$, which starts from 0 at object boundary and increases monotonically when $x$ is away from the boundary. Similarly, we can define an energy function $d(x,\Gamma_s)$ representing the distance from the skeleton. It vanishes to 0 when the pixel approaches the object skeleton.
Formally, we define the energy function of the {\em skeleton-aware} distance transform (Fig.~\ref{fig:sdt_energy}) as
\begin{equation}\label{eq:sdt}
    E_{SDT}(x) = \left(\frac{d(x,\Gamma_b)}{d(x,\Gamma_s) + d(x,\Gamma_b)}\right)^\alpha,~\alpha > 0
\end{equation}
where $\alpha$ controls the curvature of the energy surface\footnote{We add $\epsilon=10^{-6}$ to the denominator to avoid dividing by 0 for the edge case where a pixel is both instance boundary and skeleton (\ie, a one-pixel wide part).}. 
When $0<\alpha<1$, the function is concave and decreases faster when being close to the boundary, and vice versa when $\alpha>1$. 
In the ablation studies, we demonstrate various patterns of the model predictions given different $\alpha$.

Besides, since common skeletonization algorithms can be sensitive to small perturbations on the object boundary and produce unwanted branches, we smooth the masks before computing the object skeleton by Gaussian filtering and thresholding to avoid complex branches.
\input{tex-fig/fig_sdt_energy}

\bfsection{Learning Strategy} Given the ground-truth \sdt energy map, there are two ways to learn it using a CNN model. The first way is to {\em regress} the energy using $L_1$ or $L_2$ loss. In the regression mode, the output is a single-channel image. The second way is to quantize the $[0,1]$ energy space into $K$ bins and rephrase the regression task into a {\em classification} task~\cite{bai2017deep,wang2020deep}, which makes the model robust to small perturbations in the energy landscape. For the classification mode, the model output has $(K+1)$ channels with one channel representing the background region. We fix the bin size to 0.1 without tweaking, making $K=10$. Softmax is applied before calculating the cross-entropy loss. We test both learning strategies in the experiments to illustrate the optimal setting for \sdtt.

\subsection{SDT Network}\label{sec:implementation}

\bfsection{Network Architecture} Directly learning the energy function with a fully convolutional network (FCN) can be challenging. Previous approaches either first regress an easier direction field representation and then use additional layers to predict the desired target~\cite{bai2017deep}, or take the multi-task learning approach to predict additional targets at the same time~\cite{xu2017gland,yan2018deep,shen2017deepskeleton}.

\input{tex-fig/fig_pipeline.tex}
\input{tex-fig/fig_global_local}

Fortunately, with recent progress in FCN architectures, it becomes feasible to learn the target energy map in an end-to-end fashion. Specifically, in all the experiments, we use a DeepLabV3 model~\cite{chen2017deeplab} with a ResNet~\cite{he2016deep} backbone to directly learn the \sdt energy without additional targets (Fig.~\ref{fig:pipeline}, {\em Training Phase}). We also add a CoordConv~\cite{liu2018intriguing} layer before the 3rd stage in the backbone network to introduce spatial information into the segmentation model.

\bfsection{Target SDT Generation}\label{sec:g_local} There is an inconsistency problem in object skeleton generation: part of the complete instance skeleton can be different from the skeleton of the instance part (Fig.~\ref{fig:global_local}). Some objects may touch the image border due to either a restricted field of view (FoV) of the imaging devices or spatial data augmentation like the random crop. If pre-computing the skeleton, we will get {\em local skeleton} (Fig.~\ref{fig:global_local}\red{c}) for objects with missing masks due to imaging restrictions, and {\em partial skeleton} (Fig.~\ref{fig:global_local}\red{b}) due to spatial data augmentation, which causes ambiguity. Therefore we calculate the local skeleton for \sdt on-the-fly after all spatial transformations instead of pre-computing to prevent the model from hallucinating the structure of parts outside of the currently visible region. In inference, we always run predictions on the whole images to avoid inconsistent predictions. We use the skeletonization algorithm in Lee \etal~\cite{lee1994building}, which is less sensitive to small perturbations and produces skeletons with fewer branches.

\bfsection{Instance Extraction from SDT}
In the \sdt energy map, all boundary pixels share the same energy value and can be processed into segments by direct thresholding and connected component labeling, similar to DWT~\cite{bai2017deep}. However, since the prediction is never perfect, the energy values along closely touching boundaries are usually not sharp and cause split-errors when applying a higher threshold or merge-errors when applying a lower threshold.

Therefore we utilize a skeleton-aware instance extraction (Fig.~\ref{fig:pipeline}, {\em Inference Phase}) for \sdtt. Specifically, we set a threshold $\theta=0.7$ so that all pixels with the predicted energy bigger than $\theta$ are labeled as skeleton pixels. We first perform connected component labeling of the skeleton pixels to generate seeds and run the watershed algorithm on the reversed energy map using the seeds as basins (local optima) to generate the final segmentation. We also follow previous works~\cite{chen2016dcan,yan2018deep} and refine the segmentation by hole-filling and removing small spurious objects.

%% file: tex-fig/fig_sdt_energy.tex
\begin{figure}[t]
    \centering    \includegraphics[width=0.8\columnwidth]{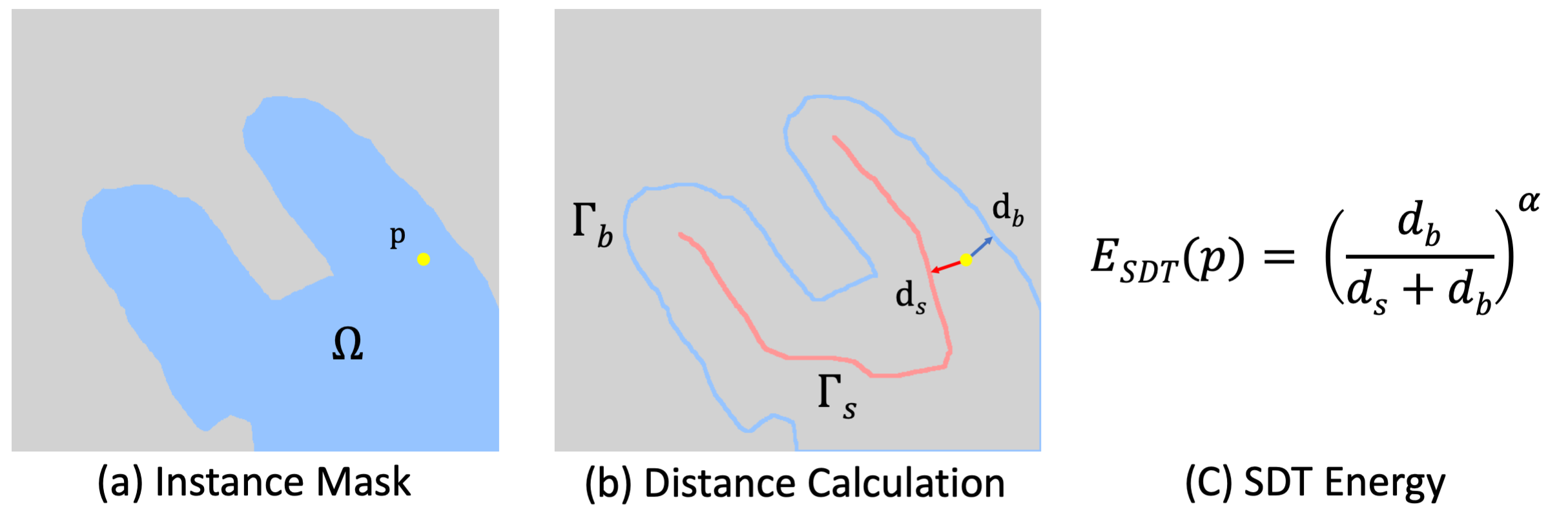}
    \caption{Illustration of the SDT energy function. ({\bf a}) Given an instance mask $\Omega$, ({\bf b}) we calculate the distances of a pixel to both the skeleton and boundary. ({\bf c}) Our energy function ensures a uniform maximum value of 1 on the skeleton and minimum value of 0 on the boundary, with a smooth interpolation in between.
    }\label{fig:sdt_energy}
\end{figure}

%% file: tex-fig/fig_pipeline.tex
\begin{figure*}[t]
    \centering
    \includegraphics[width=0.95\textwidth]{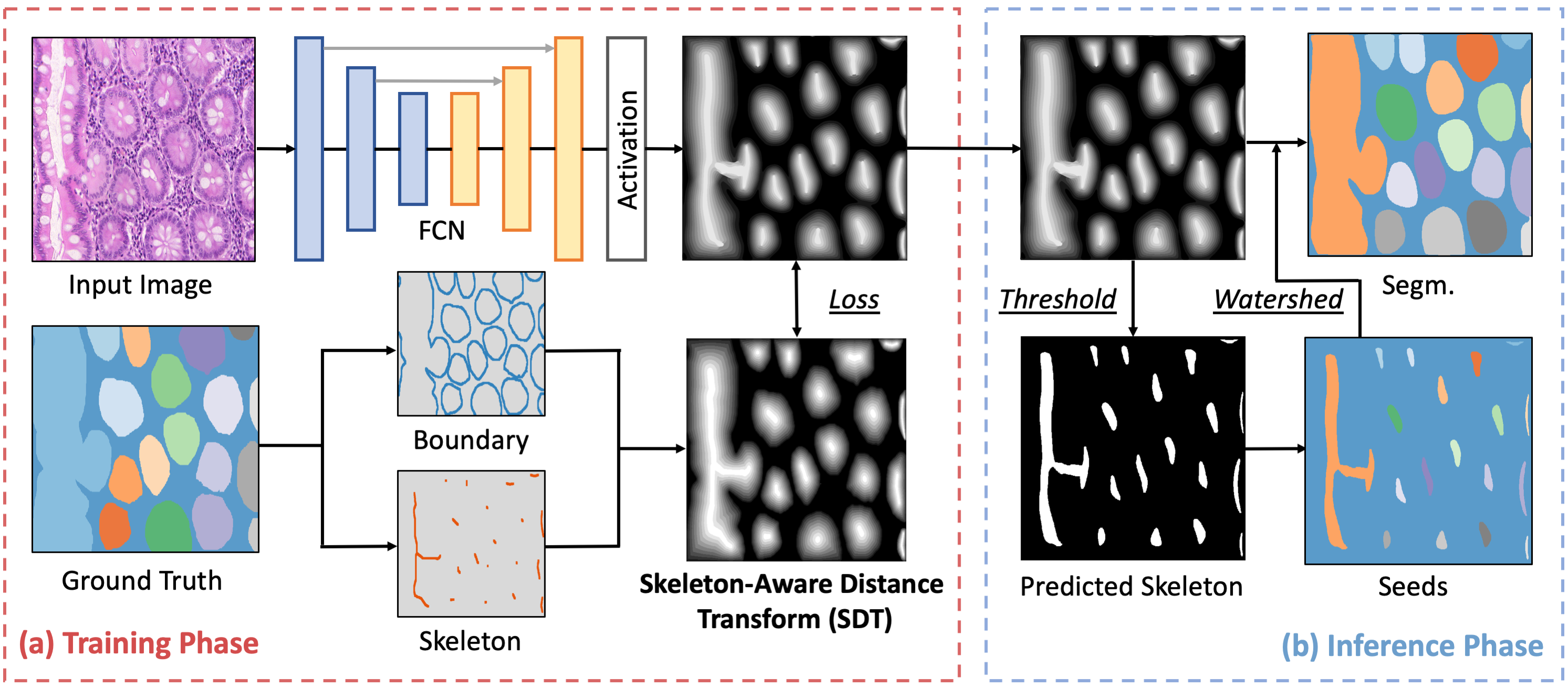}
    \caption{
    Overview of the SDT framework. {\bf(a)} {\em Training Phase}: target SDT is calculated conditioned on the distance to both the boundary and skeleton. A FCN maps the image into the energy space to minimize the loss. {\bf (b)} {\em Inference Phase}: we threshold the SDT to generate skeleton segments, which is processed into seeds with the connected component labeling. Finally, the watershed transform algorithm takes the seeds and the reversed SDT energy to yield the masks.
    }\label{fig:pipeline}
\end{figure*}

%% file: tex-fig/fig_global_local.tex
\begin{figure}[t]
    \centering    \includegraphics[width=0.75\columnwidth]{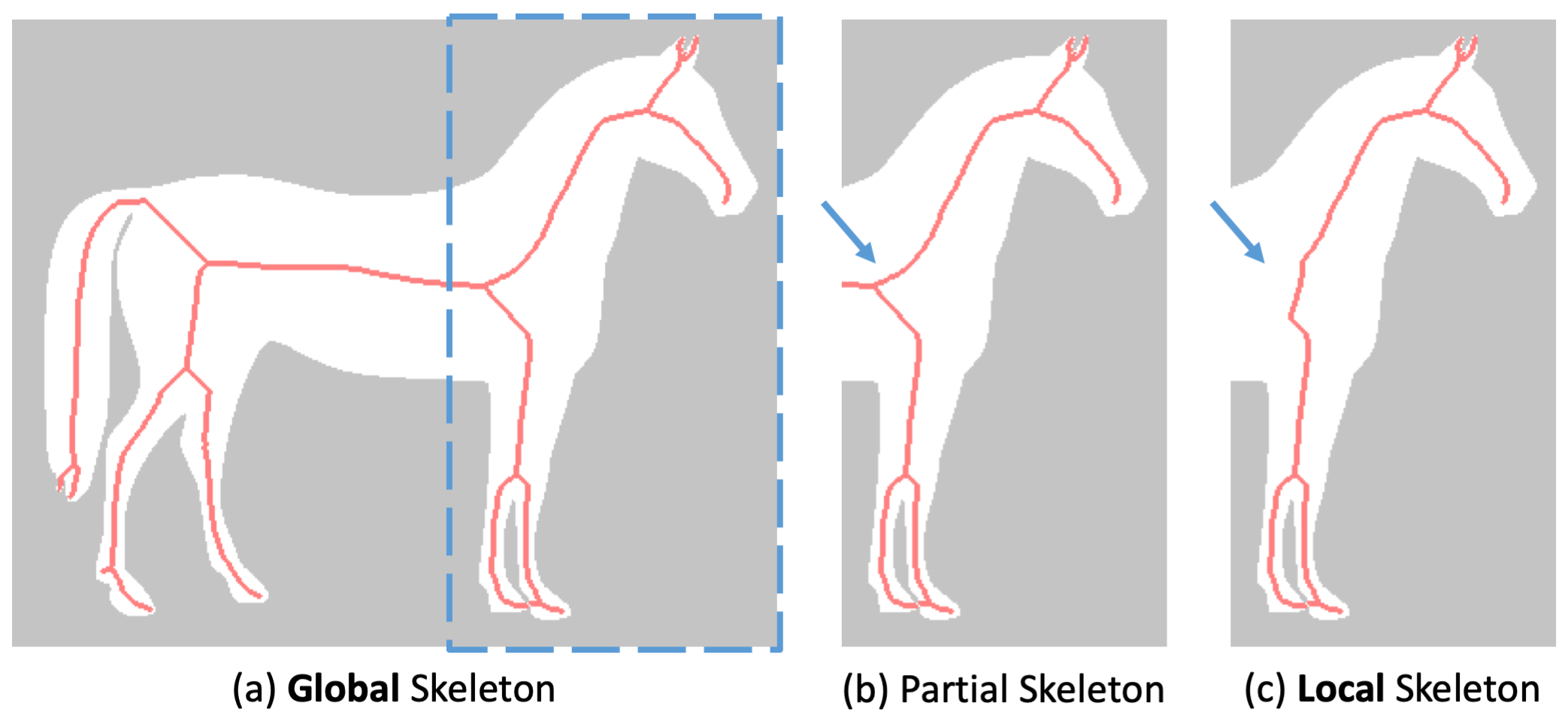}
    \caption{Skeleton generation rule. ({\bf a}) Given an instance and the {\em global} skeleton, ({\bf b}) the {\em partial} skeleton cropped from the global skeleton can be different from ({\bf c}) the {\em local} skeleton generated from the cropped mask. For \sdtt, we calculate the local skeleton to prevent the model from extrapolating the unseen parts. 
    }\label{fig:global_local}
\end{figure}

%% file: sections/4_experiment.tex
\section{Experiments}\label{sec:exp}

\input{sections/4_exp_gland.tex}
\input{sections/4_exp_ablation}


%% file: sections/4_exp_gland.tex
\subsection{Histopathology Instance Segmentation}\label{subsec:exp_gland}
Accurate instance segmentation of gland tissues in histopathology images is essential for clinical analysis, especially cancer diagnosis. The diversity of object appearance, size, and shape makes the task challenging. 

\bfsection{Dataset and Evaluation Metric}
We use the gland segmentation challenge dataset~\cite{sirinukunwattana2017gland} that contains colored light microscopy images of tissues with a wide range of histological levels from benign to malignant. There are 85 and 80 images in the training and test set, respectively, with ground truth annotations provided by pathologists. According to the challenge protocol, the test set is further divided into two splits with 60 images of normal and 20 images of abnormal tissues for evaluation. Three evaluation criteria used in the challenge include instance-level F1 score, Dice index, and Hausdorff distance, which measure the performance of object detection, segmentation, and shape similarity, respectively. For the instance-level F1 score, an IoU threshold of 0.5 is used to decide the correctness of a prediction.

\input{tex-fig/fig_gland}

\bfsection{Methods in Comparison} 
We compare \sdt with previous state-of-the-art segmentation methods, including DCAN~\cite{chen2016dcan}, multi-channel network (MCN)~\cite{xu2017gland}, shape-preserving loss (SPL)~\cite{yan2018deep} and FullNet~\cite{qu2019improving}. We also compare with suggestive annotation (SA)~\cite{yang2017suggestive}, and SA with model quantization (QSA)~\cite{xu2018quantization}, which use multiple FCN models to select informative training samples from the dataset. With the same training settings as our \sdtt, we also report the performance of skeleton with scales (SS) and traditional distance transform (DT).

\bfsection{Training and Inference}
Since the training data is relatively limited due to the challenges in collecting medical images, we apply pixel-level and spatial-level augmentations, including random brightness, contrast, rotation, crop, and elastic transformation, to alleviate overfitting. We set $\alpha=0.8$ for our \sdt in Eqn.~\ref{eq:sdt}. We use the classification learning strategy and optimize a model with 11 output channels (10 channels for energy quantized into ten bins and one channel for background). We train the model for 20k iterations with an initial learning rate of $5\times10^{-4}$ and a momentum of $0.9$. The same settings are applied to DT. At inference time, we apply argmax to get the corresponding bin index of each pixel and transform the energy value to the original data range. Finally, we apply the watershed-based instance extraction rule described in Sec.~\ref{sec:implementation}.

Specifically for SS, we set the number of output channels to two, with one channel predicting skeleton probability and the other predicting scales. Since the scales are non-negative, we add a ReLU activation for the second channel and calculate the regression loss. Masks are generated by morphological dilation. We do not quantize the scales as DT and SDT since even ground-truth scales can yield masks unaligned with the instance boundary with quantization.

\bfsection{Results} 
Our \sdt framework achieves state-of-the-art performance on 5 out of 6 evaluation metrics on the gland segmentation dataset (Table~\ref{tab:gland}). With the better distinguishability of object interior and boundary, \sdt can unambiguously separate closely touching instances (Fig.~\ref{fig:gland}, first two rows), performs better than previous methods using object boundary representations~\cite{chen2016dcan,yan2018deep}. Besides, under the Hausdorff distance for evaluating shape-similarity between ground-truth and predicted masks, our \sdt reports an average score of 44.82 across two test splits, which improves the previous state-of-the-art approach (\ie, FullNet with an average score of 50.15) by $10.6\%$. 
We also notice the different sensitivities of the three evaluation metrics. 
Taking the instance {\bf D} (Fig.~\ref{fig:gland}, 3rd row) as an example:  both \sdt and FullNet~\cite{qu2019improving} have 1.0 F1-score (IoU threshold 0.5) for the correct detection; \sdt has a slightly higher Dice Index (0.956 vs.  0.931) for better pixel-level classification;  and our \sdt has significantly lower Hausdorff distance (24.41 vs. 48.81) as SDT yields a mask with much more accurate morphology.



%% file: tex-fig/fig_gland.tex
\begin{figure*}[t]
    \centering
    \includegraphics[width=\textwidth]{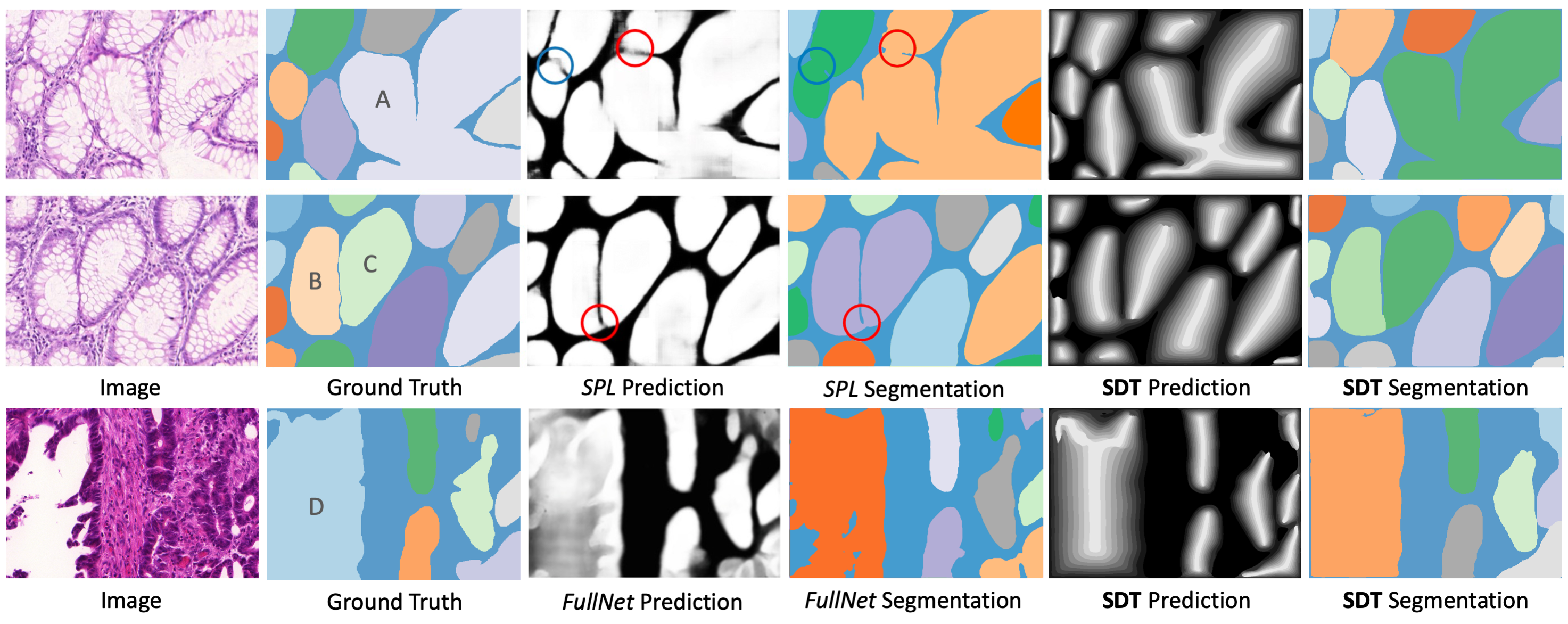}
    \caption{
    Visual comparison on histopathology image segmentation. (First 2 rows) Compared with shape-preserving loss (SPL)~\cite{yan2018deep}, our \sdt unambiguously separates closely touching objects while preserving the structure of complicated masks. (The 3rd row) Compared with FullNet~\cite{qu2019improving}, our model infers the \sdt energy of instance masks from a global structure perspective instead of boundary that relies on relatively local predictions, which produces high-quality masks.
    }\label{fig:gland}
\end{figure*}

%% file: sections/4_exp_ablation.tex
\subsection{Ablation Studies}\label{sec:ablation}


\bfsection{Loss Function} We compare the regression mode using L1 and L2 losses with the classification mode using cross-entropy loss. There is a separate channel for background under the classification mode where the energy values are quantized into bins. However, for regression mode, if the background value is 0, we need to use a threshold $\tau>0$ to decide the foreground region, which results in shrank masks. To separate the background region from the foreground objects, we assign an energy value of $-b$ to the background pixels $(b \geq 0)$. To facilitate the regression, given the predicted value $\hat{y}_i$ for pixel $i$, we apply a sigmoid function ($\sigma$) and affine transformation so that
$\hat{y}_i^\prime = (1+b)\cdot \sigma(\hat{y}_i) - b$ has a range of $(-b,1)$. We set $b=0.1$ for the experiments.
We show that under the same settings, the model trained with quantized energy reports the best results (Table~\ref{tab:gland_ablation}). We also notice that the model trained with $L_1$ loss produces a much sharper energy surface than the model trained with $L_2$ loss, which is expected.

\input{tex-tab/tab_two}

\bfsection{Curvature} We also compare different $\alpha$ in Eqn.~\ref{eq:sdt} that controls the curvature of the energy landscape. Table~\ref{tab:gland_ablation} shows that $\alpha=0.8$ achieves the best overall performance, which is slightly better than $\alpha=1.0$. Decreasing $\alpha$ to 0.6 introduces more merges and make the results worse.

\bfsection{Global/Local Skeleton} In Sec.~\ref{sec:implementation} we show the inconsistency problem of global and local skeletons. In this study, we set $\alpha=0.8$ and let the model learn the pre-computed \sdt energy for the training set. The results show that pre-computed \sdt significantly degrades performance (Table~\ref{tab:gland_ablation}). We argue this is because pre-computed energy not only introduces inconsistency for instances touching the image border but also restricts the diversity of \sdt energy maps.

%% file: tex-tab/tab_two.tex
\begin{table}[t]
\begin{minipage}[t]{0.51\textwidth}\vspace{0pt}
\centering
    \resizebox{\columnwidth}{!}{
    \begin{tabular}{lcccccc}
    \toprule
    \multirow{2}{*}{Method} &  \multicolumn{2}{c}{F1 Score $\uparrow$} & \multicolumn{2}{c}{Dice Index $\uparrow$} &
    \multicolumn{2}{c}{Hausdorff $\downarrow$}\\
    \cmidrule{2-7}
    {}  & Part A & Part B & Part A & Part B & Part A & Part B\\
    \midrule
    DCAN~\cite{chen2016dcan}   
    & 0.912 & 0.716 & 0.897 & 0.781 & 45.42 & 160.35\\
    MCN~\cite{xu2017gland}  
    & 0.893 & 0.843 & 0.908 & 0.833 & 44.13 & 116.82\\
    SPL~\cite{yan2018deep}   & 0.924  & 0.844  & 0.902  & 0.840  & 49.88 & 106.08\\
    SA~\cite{yang2017suggestive} & 0.921 & 0.855 & 0.904 & 0.858 & 44.74 & 96.98\\
    FullNet~\cite{qu2019improving} & 0.924  & 0.853  & 0.914  & 0.856 & 37.28 & 88.75\\
    QSA~\cite{xu2018quantization} & 0.930 & 0.862 & 0.914 & {\bf 0.859} & 41.78 & 97.39\\
    \midrule
    SS & 0.872 & 0.765 & 0.853 & 0.797 & 54.86 & 116.33\\
    DT & 0.918 & 0.846 & 0.896 & 0.848 & 41.84 & 90.86\\
    \textbf{SDT} & {\bf 0.931} & {\bf 0.866} & {\bf 0.919} 
    & 0.851 & {\bf 32.29} & {\bf 82.40}\\
    \bottomrule
    \end{tabular}}
    \vspace{0.05in}
    \captionof{table}{Comparison with existing methods on the gland segmentation. Our \sdt achieves better or on par F1 score and Dice Index, and significantly better Hausdorff distance for evaluating {\em shape similarity}. DT and SS represent distance transform and skeleton with scales.}\label{tab:gland}
\end{minipage}
~
\begin{minipage}[t]{0.48\textwidth}\vspace{0pt}
\centering
    \resizebox{\columnwidth}{!}{
    \begin{tabular}{lcccccc}
    \toprule
    \multirow{2}{*}{Setting} &  \multicolumn{2}{c}{F1 Score $\uparrow$} & \multicolumn{2}{c}{Dice Index $\uparrow$} &
    \multicolumn{2}{c}{Hausdorff $\downarrow$}\\
    \cmidrule{2-7}
    {}  & Part A & Part B & Part A & Part B & Part A & Part B\\
    \midrule
    {\em Loss} \\
    L1 & 0.916 & 0.842 & 0.903 & 0.850 & 39.76 & 94.83\\
    L2 & 0.896 & 0.833 & 0.885 & 0.837 & 49.11 & 110.24\\
    CE & 0.931 & 0.866 & 0.919 & 0.851 & 32.29 & 82.40\\
    \midrule
    {\em Curvature} \\
    $\alpha=0.6$ & 0.912 & 0.845 & 0.914 & 0.855 & 36.25 & 91.24 \\
    $\alpha=0.8$ & 0.931 & 0.866 & 0.919 & 0.851 & 32.29 & 82.40 \\
    $\alpha=1.0$ & 0.926 & 0.858 & 0.907 & 0.849 & 35.73 & 86.73 \\
    \midrule
    {\em Skeleton} \\
    Partial & 0.899 & 0.831 & 0.896 & 0.837 & 47.50 & 105.19\\
    Local & 0.931 & 0.866 & 0.919 & 0.851 & 32.29 & 82.40 \\
    \bottomrule
    \end{tabular}}
    \vspace{0.05in}
    \captionof{table}{Ablations studies on the gland dataset. The results suggest that the model trained with cross-entropy loss with $\alpha=0.8$ and local skeleton generation achieves the best performance.}\label{tab:gland_ablation}
\end{minipage}
\end{table}